\documentclass[10pt,a4paper,twocolumn]{article}
\usepackage[T1]{fontenc}
\usepackage[utf8]{inputenc}
\usepackage{mathptmx}
\usepackage{amsmath,amssymb,bm}
\usepackage[a4paper,textwidth=170mm,textheight=235mm,centering]{geometry}
\usepackage{graphicx}
\usepackage{array,booktabs}
\usepackage[font=small,textfont=it,tableposition=top]{caption}
\usepackage{tikz}
\usepackage{xurl}
\usepackage[hidelinks]{hyperref}
\newcommand{\sdk}{GigaEdge Enhance}
\newcommand{\enEight}{Enhance 8 kHz INT8}
\newcommand{\enDisc}{Enhance 48 kHz Discriminative}
\newcommand{\enHybrid}{Enhance 48 kHz Hybrid}
\newcommand{\dpdf}{DPDFNet2 48 kHz HR}
\newcommand{\paperkeywords}{speech enhancement, streaming inference, cross-platform deployment, temporal frame grouping, computational efficiency}

\title{GigaEdge Enhance: An SDK for Practical Streaming Speech Enhancement}
\author{%
  \normalsize Artem Sokolov,
  Artem Khaidarov,
  Yana Lysova,
  Alexander Gladkikh,
  Vitaliy Yulis\\[5pt]
  \small SaluteDevices, Russia\\[4pt]
  \footnotesize\texttt{artsokol87@gmail.com, akhdrv@gmail.com}\\
  \footnotesize\texttt{yana.ml.engineer@gmail.com, gladkihaa38@gmail.com, vitaly.yulis@gmail.com}%
}
\date{}
\hypersetup{
  pdftitle={GigaEdge Enhance: An SDK for Practical Streaming Speech Enhancement},
  pdfauthor={Artem Sokolov, Artem Khaidarov, Yana Lysova, Alexander Gladkikh, Vitaliy Yulis},
  pdfkeywords={speech enhancement, streaming inference, cross-platform deployment, temporal frame grouping, computational efficiency}
}

\begin{document}
  \maketitle
  \begin{abstract}
  We present GigaEdge Enhance, an open-source SDK for streaming speech enhancement on CPU servers, mobile devices, and in browsers. Applications can build enhancement channels from neural denoising, normalization and dynamics blocks, sample-rate conversion, and acoustic echo cancellation where supported. These stages can be combined, while normalization can also run without a neural model. The release includes an 8 kHz INT8 denoiser and discriminative and hybrid 48 kHz FP32 models. A shared C++ core supports ONNX Runtime and TensorFlow Lite through native interfaces, Android and iOS SDKs, and WebAssembly integration. Independent stream contexts share model resources, and consecutive frames from one stream can be processed in a single inference call. We evaluate the released models on EARS--WHAM v2 and compare their quality and processing costs with RNNoise and DPDFNet2. All tested native and Web configurations process files faster than real time, although larger chunks require more time to collect input. We also measure how many streams a shared model can process on a CPU. Our contribution is a configurable deployment system and an experimental study of how model choice, chunk size, and execution environment affect its use. All reported SDK timing uses ONNX Runtime.
  \end{abstract}
  \par\medskip
  
  \section{Introduction}
  Speech enhancement is an important component of voice communication and speech interfaces. Neural approaches can improve the quality of noisy recordings, while efficient implementations make local processing practical~\cite{rnnoise,dtln,dfn2}. However, a model used in a live application must accept successive audio chunks and preserve its state while working with an audio callback that may use a different block size. These requirements also apply to the signal-processing stages outside the neural graph, since the application depends on the complete audio path.
  
  In this paper, we describe \sdk, a software development kit for building real-time audio enhancement pipelines on CPU servers and end-user devices. In the SDK, an enhancement channel is a configurable processing pipeline. Applications can use normalization alone, neural denoising alone, or both, and add acoustic echo cancellation where the platform API supports it. The normalization stage can itself be built from selectable signal-processing blocks.

  The shared core supports ONNX Runtime and TensorFlow Lite through native, mobile, and browser interfaces. We study this system, the quality of its released models, and its processing costs in different execution environments. All reported SDK performance measurements use ONNX Runtime, while the external RNNoise reference uses its native C kernels.
  
  The system has three main components. A shared C++ core handles sample-rate conversion, spectral processing, and model state for each stream, with C and C++ interfaces for its audio-processing modules. Platform wrappers connect this core to Android, iOS, and Web Audio applications and provide worker-based execution for live audio. The open-source release also supplies three models for 8 kHz INT8 and 48 kHz FP32 processing, although training, quantization, and conversion tools are outside this SDK release. We study how chunk size and execution environment affect processing cost, measure concurrent native throughput, and evaluate all three models through the SDK CLI in separate bandwidth conditions.
  
  We developed the SDK for videoconferencing across native platforms and browsers, where devices differ in compute resources and microphone quality. Enhancement must share resources with video encoding and decoding, including demanding codecs, while keeping audio free from interruptions and other processing artifacts. These requirements shaped the models' quality/computation trade-offs. A high quality score alone does not show whether a model fits such an application, because full processing cost, buffering, and concurrent-stream capacity impose different limits. We therefore examine three questions: how the released models balance predicted quality and computation; how chunk size affects native and Web processing; and how many independent streams can share a model while each meets a processing-rate requirement.

  Our contribution is a deployment design and an experimental study that links model bandwidth, the number of consecutive frames in each inference call, and the number of streams sharing model resources. We combine these choices in configurable enhancement channels built on one C++ core. The experiments show that larger chunks can provide more throughput than adding CLI processes, and that the model with the lowest arithmetic cost does not necessarily offer the best balance of quality and measured capacity. We compare these choices with RNNoise and sherpa-onnx/DPDFNet2, stating which operations and processing stages are counted. Recurrent-state caching, model sharing, ring buffers, and worker threads are established methods; our system contribution is their integration and the study of the resulting deployment choices. The detailed architecture and training objectives of the 48 kHz hybrid model are outside this study's scope.
  
  \section{Related work}
  \subsection{Research software and efficient enhancement}
  Asteroid provides modular architectures and reproducible separation recipes~\cite{asteroid}, while ESPnet-SE combines enhancement and separation with ASR~\cite{espnetse}. Both show how research software can be supported by benchmark experiments. We focus on streaming in applications, where model execution must work together with signal processing and host audio interfaces.

  RNNoise combines neural estimation with DSP and provides an inference library, pretrained weights, feature-generation tools, and a training procedure~\cite{rnnoise,rnnoiserepo}. It can also run in browsers: Jitsi's RNNoise WebAssembly wrapper includes a synchronous 0.2 build for AudioWorklet use and an older asynchronous build~\cite{rnnoisewasm}. We use the regular RNNoise 0.2 model as a strong native efficiency reference, while its WebAssembly wrapper is outside our measurements.

  DTLN provides streaming inference with externally managed recurrent state and ONNX/TensorFlow Lite conversion~\cite{dtln,dtlnrepo}. DeepFilterNet2 targets efficient full-band processing~\cite{dfn2}, and later DeepFilterNet work provides real-time enhancement with a public Rust implementation and deployment tools~\cite{dfn3,dfnrepo}. GTCRN targets very low computational requirements~\cite{gtcrn}. DPDFNet extends DeepFilterNet2 with dual-path recurrence~\cite{dpdfnet}; its implementation includes 8, 16, and 48 kHz models, stateful streaming, and ONNX/TensorFlow Lite inference~\cite{dpdfnetrepo}. These systems provide a basis for application deployment through efficient CPU inference, recurrent-state caching, model export, and support for different sampling rates.

  \subsection{Integrated audio processing and platform runtimes}
  WebRTC's Audio Processing Module combines noise suppression, acoustic echo cancellation, and automatic gain control~\cite{webrtcapm}. SpeexDSP also provides established audio-processing components, including a preprocessor with noise suppression, residual echo suppression, gain control, and VAD~\cite{speexdsp}. Our SDK follows this approach of combining processing stages and uses SpeexDSP for echo cancellation.

  sherpa-onnx is a relevant deployment reference because it offers C/C++ and other language interfaces for native, Android, iOS, and WebAssembly environments, including streaming enhancement with GTCRN and DPDFNet~\cite{sherpa,sherpaenhancement}. It therefore already supports cross-platform neural enhancement. We use its DPDFNet2 48 kHz HR configuration as an external reference for quality and CPU capacity, with the shared-model adaptation described in Section~\ref{sec:eval}. Table~\ref{tab:related-systems} compares the focus of these systems using their papers and official documentation as of October 2026.

  \begin{table*}[t]
  \centering
  \small
  \setlength{\tabcolsep}{3pt}
  \caption{Related software and the focus of this work. The table summarizes documented functions rather than all available features, and does not rank the systems by measured quality or speed.}
  \label{tab:related-systems}
  \renewcommand{\arraystretch}{1.08}
  \begin{tabular}{@{}>{\raggedright\arraybackslash}p{0.18\textwidth}>{\raggedright\arraybackslash}p{0.34\textwidth}>{\raggedright\arraybackslash}p{0.42\textwidth}@{}}
  \toprule
  System & Documented focus & Relation to \sdk \\
  \midrule
  \textbf{Asteroid; ESPnet-SE}~\cite{asteroid,espnetse} & Research architectures, recipes, and evaluation; ASR integration in ESPnet-SE. & We study the deployment pipeline, host audio integration, and concurrent CPU workloads. \\
  \addlinespace[3pt]\midrule\addlinespace[3pt]
  \textbf{RNNoise}~\cite{rnnoise,rnnoiserepo,rnnoisewasm} & DSP/neural enhancement, pretrained weights and training tools; shared weights with separate states; third-party WebAssembly integration. & Our native efficiency reference, evaluated at 48 kHz and after output downsampling to 8 kHz. We study model selection and temporal frame grouping in configurable channels with different sampling rates. \\
  \addlinespace[3pt]\midrule\addlinespace[3pt]
  \textbf{DTLN; DeepFilterNet}~\cite{dtln,dfn3} & Efficient enhancement models with reusable inference and training implementations. & We package model choices together with stream management, platform wrappers, and configurable audio processing. \\
  \addlinespace[3pt]\midrule\addlinespace[3pt]
  \textbf{GTCRN; DPDFNet}~\cite{gtcrn,dpdfnet,dpdfnetrepo} & Low-cost enhancement architectures; DPDFNet also supplies multirate streaming and deployment formats. & Our experiments address SDK-level chunk costs and shared-model capacity for the released model configurations. \\
  \addlinespace[3pt]\midrule\addlinespace[3pt]
  \textbf{WebRTC APM; SpeexDSP}~\cite{webrtcapm,speexdsp} & Established integrated audio processing for communication applications. & We combine neural model selection with these types of application components; our AEC reuses SpeexDSP. \\
  \addlinespace[3pt]\midrule\addlinespace[3pt]
  \textbf{sherpa-onnx}~\cite{sherpa,sherpaenhancement} & Cross-platform speech runtime; the inspected upstream streaming denoisers combine a model and one stream state. & We expose model/context separation in the SDK, temporal frame grouping, configurable level control, native/Web AEC, and independent input/output rates. The capacity reference uses a shared-model adaptation of sherpa-onnx. \\
  \addlinespace[3pt]\midrule\addlinespace[3pt]
  \textbf{\sdk\ (this work)} & Configurable DSP/neural enhancement channels with one C++ core, ONNX/TensorFlow Lite adapters, and native/mobile/Web interfaces. & Processing stages can run alone or together, with selectable normalization blocks. We evaluate the released neural configurations. \\
  \bottomrule
  \end{tabular}
  \end{table*}

  \subsection{Positioning and streaming validation}
  Compared with sherpa-onnx, we focus on how the enhancement pipeline is organized. In the GTCRN and DPDFNet streaming implementations that we inspected, each denoiser owns a model and one stream's state, and processes one spectral frame per neural call~\cite{sherpastreaming}. In \sdk, independent stream contexts share a model instance and its ONNX session, so a new stream does not require a separate inference session. Our released ONNX models also accept several consecutive frames in one call, which allows larger chunks to change how inference is performed as well as how audio is passed to the SDK. Sections~\ref{sec:eval} and~\ref{sec:results} examine chunk-size effects and shared-model capacity. For the capacity comparison, we adapt sherpa-onnx to share a model and session across independent contexts as well. Both measured configurations therefore share model resources, and session duplication cannot explain their capacity difference. This API comparison applies to the inspected enhancement implementations rather than all speech-processing APIs in sherpa-onnx.

  RNNoise already separates each stream's denoising state from shared weights. Its API provides independent \texttt{DenoiseState} objects that can use the same loaded \texttt{RNNModel}; states using the default model also refer to the compiled weight arrays~\cite{rnnoiseapi}. Model/state separation is therefore an established method. We combine it with model and backend selection, temporal frame grouping, independent I/O rates, and configurable DSP channels.

  A second difference is the processing chain exposed to applications. The public sherpa-onnx enhancement API configures the model and inference settings, but does not expose a playback-reference AEC input or configurable loudness-control stages~\cite{sherpaseapi}. Its streaming adapter supports input resampling and returns audio at the model rate~\cite{sherpastft}. In \sdk, applications can combine neural denoising with selected normalization blocks and, through native and Web interfaces, SpeexDSP AEC. They can also choose the input and output rates independently for neural enhancement. Normalization and denoising can run separately or together, and applications can select the blocks within the normalization stage. Table~\ref{tab:platforms} shows the AEC limits of the mobile interfaces.

  Temporal caching is an established way to avoid recomputing history, as used, for example, in streaming Wave-U-Net~\cite{waveunet}. Each stream in our implementation also retains its resamplers, spectral overlap, output buffers, and statistics. Regression tests in the underlying product compare sequential processing with larger and mixed-size chunks, checking startup alignment and finalization within numerical tolerances.

  \section{System design}
  \subsection{Enhancement channels and processing layers}
  The SDK and the three described models are publicly available in our open-source GitHub repository.\footnote{\url{https://github.com/salute-developers/giga-edge-enhance}}
  
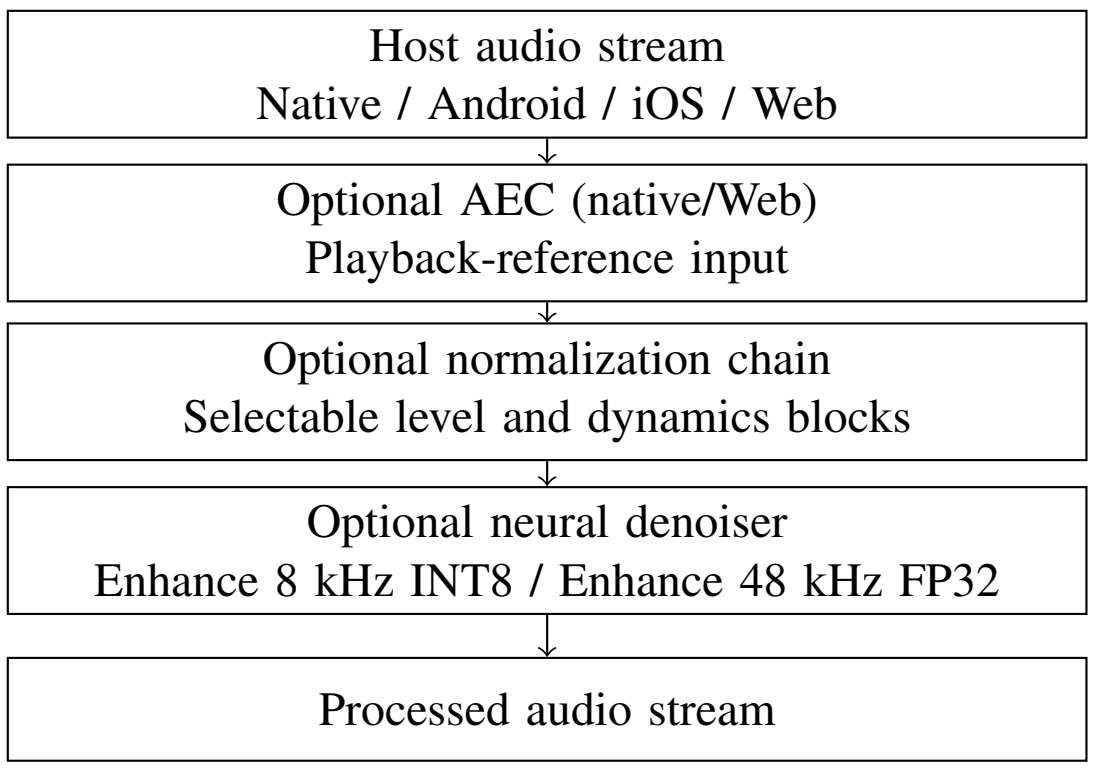
\begin{figure}[t]
  \centering
  \begin{tikzpicture}[every node/.style={font=\small,align=center},box/.style={draw,text width=6.9cm,minimum height=0.68cm}]
  \node[box] (a) at (0,0) {Host audio stream\\Native / Android / iOS / Web};
  \node[box] (b) at (0,-1.05) {Optional AEC (native/Web)\\Playback-reference input};
  \node[box] (c) at (0,-2.10) {Optional normalization chain\\Selectable level and dynamics blocks};
  \node[box] (d) at (0,-3.15) {Optional neural denoiser\\Enhance 8 kHz INT8 / Enhance 48 kHz FP32};
  \node[box] (e) at (0,-4.20) {Processed audio stream};
  \draw[->] (a)--(b);
  \draw[->] (b)--(c);
  \draw[->] (c)--(d);
  \draw[->] (d)--(e);
  \end{tikzpicture}
  \caption{An example enhancement channel. Stages can be omitted, while normalization can itself combine selected DSP blocks. This is one possible order; available stages and controls depend on the platform interface. The neural stage uses ONNX Runtime or TensorFlow Lite according to model and platform support (Table~\ref{tab:platforms}).}
  \label{fig:system}
  \end{figure}

  An enhancement channel is an ordered processing pipeline applied to an audio stream. Here, channel refers to the pipeline rather than waveform channel count, such as mono or stereo. Applications select the DSP and neural stages and then choose the blocks within the normalization chain. A normalization-only channel runs without loading a neural model, while a denoising-only channel omits normalization. A combined channel uses both, with the available stages and composition controls defined by the platform API. Figure~\ref{fig:system} shows an example channel built from the shared C++ components. The library provides DSP implementations and model-specific inference adapters, while platform bindings connect them to host audio and manage resource loading, initialization, and cleanup. This allows applications to reuse the same processing code on different hosts and in different channel configurations.
  
  The released discriminative configurations use a 32 ms analysis window and a 16 ms hop at model rates of 8 or 48 kHz. The corresponding window/hop lengths are 256/128 and 1536/768 samples. Processing chunks have durations $T_{\mathrm{chunk}}=m\cdot16$ ms, where $m$ is a positive integer within the implementation's size limit. This gives $128m$ samples at 8 kHz or $768m$ samples at 48 kHz. The ONNX adapters pass the consecutive spectral frames of a chunk together along the model's temporal axis, so one inference call processes several frames from the same stream while keeping that stream's own recurrent state. The chunk-size rule applies after resampling and buffering at the model input; host audio callbacks may use different block sizes.
  
  The built-in resampler allows input and output sampling rates to be set independently of the model rate. When creating a stream context, the application specifies both rates, and the SDK converts incoming audio to the model rate before enhancement and the result to the requested output rate. Since each context retains its resampler state, applications do not need external sample-rate conversion. A higher output rate changes the delivered format but does not extend the bandwidth processed by the model.
  
  \subsection{Supplied models}
  The release includes three configurations: \enEight, \enDisc, and \enHybrid. The first uses INT8 quantization, while the two 48 kHz ONNX models use FP32. Table~\ref{tab:platforms} lists backend support. We developed these models for devices and platforms with different compute budgets, so both enhancement quality and processing cost were design requirements. We evaluate them as model choices within the same SDK.
  
  \enHybrid\ combines a pretrained discriminative enhancer with an additional generative component trained using a generative adversarial network (GAN) procedure. During this training, the discriminative model's weights remain frozen. The term hybrid refers to this combination, which we evaluate here as a released deployment configuration. The discriminative models are based on a modified FullSubNet architecture~\cite{fullsubnet}. In this paper, we focus on the SDK and the interfaces used to run the released models in applications.
  
  \subsection{Stream-specific state}
  For the neural denoising stage and its associated resampling and spectral processing, let $x_{j,k}$ denote the input to stream $j$ at step $k$. We express this stateful operation as
  \begin{equation}
   (\hat{s}_{j,k},z_{j,k+1})=\mathcal{P}_{\theta,c}(x_{j,k},z_{j,k}),
  \end{equation}
  where $\theta$ denotes network parameters, $c$ the processing configuration, and $z$ the state updated during processing. This state includes recurrent tensors, convolutional histories, previous input and spectral frames, resamplers, and running statistics. Each stream context stores its own state, while the model adapter owns the shared inference session, so adding a context does not create another session.
  
  This separation prevents streams from changing each other's history. However, calls on the same context must not run concurrently, and sharing a model does not imply a proportional increase in throughput.
  
  \section{Application integration}
  \subsection{Platform capabilities}
  Table~\ref{tab:platforms} summarizes model/backend support and the functions exposed by each platform interface. The shared core supports ONNX Runtime and TensorFlow Lite, while model packaging and echo-reference access differ across platforms. All reported SDK timing uses ONNX Runtime.

  Model adapters support either a predicted mask applied to the noisy spectrum or an enhanced complex spectrum passed directly to synthesis.

  \begin{table*}[t]
  \centering
  \small
  \setlength{\tabcolsep}{5pt}
  \caption{Platform and model matrix. ORT: ONNX Runtime; TL: TensorFlow Lite. A dash means that the interface does not include the model or expose the function. Backend support and the precision of a released model are separate properties.}
  \label{tab:platforms}
  \begin{tabular}{lcccc}
  \toprule
  Model or capability & C/C++ & Web & Android & iOS \\
  \midrule
  \enEight & ORT & ORT/WASM & --- & ORT \\
  \enDisc & ORT, TL & ORT/WASM & TL & ORT \\
  \enHybrid & ORT, TL & ORT/WASM & TL & ORT \\
  \midrule
  Neural noise suppression & Yes & Yes & Yes & Yes \\
  Normalization & Yes & Yes & Yes & Yes \\
  Input/output resampling & Yes & Yes & Yes & Yes \\
  AEC with echo-reference input & Yes & Yes & --- & --- \\
  \bottomrule
  \end{tabular}
  \end{table*}

  \subsection{Audio callbacks and workers}
  Android and iOS wrappers use ring buffers and a worker to handle differences between host audio blocks and model chunks. In the browser, an AudioWorklet exchanges audio with a worker through messages or shared memory; the latter requires cross-origin isolation.

  \subsection{Normalization and acoustic echo cancellation}
  Normalization is a configurable chain of level-control and dynamics blocks. The C++ DSP interface adds these blocks to an ordered sequence and allows each block's parameters to be set independently. For example, loudness control can be followed by a limiter or by a compressor and a limiter. The chain can process audio alone or together with neural denoising. These established tools help manage signal levels from microphones of different quality.
  
  \textbf{Loudness control.} This block adjusts the signal towards a target level using smoothed gain and a maximum amplification limit. Its optional speech-aware mode smooths RNNoise VAD estimates and updates loudness statistics during detected speech~\cite{rnnoise,rnnoiserepo}. Normalization can also run without VAD.

  \textbf{Automatic gain control.} This block adjusts gain towards a target signal power.

  \textbf{Peak-based range control.} This block tracks the signal envelope and adjusts gain towards a target peak level, with configurable attack, release, and hold times.

  \textbf{Dynamic range compressor.} This block reduces gain above a threshold according to a configurable compression ratio, reducing the difference between louder and quieter passages.

  \textbf{Noise gate.} This block reduces the level of signal segments below a threshold. Attack, release, and hold settings control its transitions.

  \textbf{Limiter.} This block reduces gain at high signal levels to limit output peaks according to a configured threshold.

  \textbf{Acoustic echo cancellation.} This block uses SpeexDSP~\cite{speexdsp} version 1.2.1 and a playback-reference signal to reduce playback-related echo in the microphone signal. The native and Web interfaces expose this reference input, while the Android and iOS feature APIs do not. Standalone normalization and AEC are SDK capabilities, but we do not evaluate them as enhancement methods in this study.

  \subsection{Buffer adaptation}
  The Web wrapper can increase output buffering after repeated refill events. This provides an application-level balance between playback continuity and buffering delay; the file-processing benchmarks evaluate computation and do not compare buffering policies.

  \section{Experimental setup}
  \label{sec:eval}
  We study three deployment choices: the neural model, chunk size, and concurrent stream count. We measure quality and processing cost within the SDK and compare the results with regular RNNoise 0.2 and the released DPDFNet2 checkpoint in its adapted sherpa-onnx deployment. Both external models operate at 48 kHz, with an additional quality condition for RNNoise output downsampled to 8 kHz. Normalization blocks and acoustic echo cancellation are available for building channels but are outside the experiments. SDK and sherpa-onnx performance tests use ONNX Runtime, while RNNoise uses the unmodified native library and its own inference kernels. The file-processing tests measure computation, rather than end-to-end latency or playback interruptions.
  
  \subsection{Quality evaluation}
  We evaluated the discriminative and hybrid 48 kHz ONNX models on EARS--WHAM v2~\cite{ears,wham,earsbenchmark} through the SDK's native CLI with 16 ms chunks. The benchmark contains 886 matched noisy/clean file pairs, which are also used in the separate 8 kHz condition. Input files for the two 48 kHz SDK models were sampled at the model rate, so no resampling was needed. Normalization and acoustic echo cancellation were disabled.
  
  The three released Enhance models were trained on internal datasets without EARS recordings, WHAM! recordings, or EARS--WHAM v2 mixtures. EARS--WHAM v2 is therefore an external quality benchmark for Enhance.

  Since the released checkpoints have different training histories, their quality can depend on training data, augmentation, and objectives as well as architecture. Our comparison describes deployment choices rather than isolating architecture or training method under a common training setup.
  
  All models in the 48 kHz comparison, including RNNoise and DPDFNet2, use the same wideband PESQ protocol (PESQ-WB)~\cite{pesqwb}. We evaluate the 8 kHz INT8 model on noisy audio downsampled to 8 kHz, using narrowband PESQ (PESQ-NB)~\cite{pesq}. For an additional 8 kHz output reference, RNNoise processes the original 48 kHz noisy recordings, after which its outputs are downsampled to 8 kHz and compared with the corresponding clean references using PESQ-NB. This compares two ways to produce 8 kHz audio, although RNNoise has access to the original full-band input. It does not test RNNoise on input already restricted to 8 kHz, and we report no performance measurement for this output-downsampling path. Each model is compared with noisy input at the corresponding sampling rate. PESQ-NB and PESQ-WB scores are kept separate because they cannot be used to rank models directly across bandwidths. Any sample-rate preparation within the metric implementations is separate from SDK processing.

  ESTOI estimates speech intelligibility~\cite{estoi}, PESQ assesses perceptual degradation against a clean reference, and SI-SDR measures scale-invariant reconstruction of that reference~\cite{sisdr}. For the 48 kHz condition, we also use three families of non-intrusive MOS predictors, which estimate human quality ratings from the evaluated waveform alone. The predictor families serve different purposes, as described below.

  \textbf{DNSMOS: overall quality and the noise/speech balance.} DNSMOS P.808 predicts overall recording quality and is our main overall measure within this family~\cite{dnsmos,dnsmosrepo}. The separate P.835 predictor estimates speech quality (SIG), background-noise quality (BAK), and overall quality (OVRL)~\cite{dnsmosp835}. These outputs help show whether a gain comes from a better predicted background or better predicted speech. BAK describes perceived background quality rather than noise attenuation in dB, and P.808 and P.835 OVRL are separate predictions.

  \textbf{SIGMOS: a broader view of signal impairments.} SIGMOS predicts P.804-based speech-signal quality (SIG), overall quality (OVRL), and five dimensions: spectral/timbral coloration (COL), interruptions or discontinuities (DISC), loudness (LOUD), noise (NOISE), and reverberation (REVERB)~\cite{sigmos,sigmosimplementation}. We use these dimensions to examine possible effects of enhancement beyond its noise-related score.

  \textbf{NISQA: communication quality and distortion.} NISQA was developed for transmitted speech, including communication-network impairments~\cite{nisqa}. It predicts overall MOS and four dimensions: Noisiness, Coloration, Discontinuity, and Loudness. Along with another overall-quality estimate, it helps identify predicted timbral changes, dropouts, and unsuitable signal levels. The reported NISQA model has no separate reverberation or speech-signal-quality output.

  Higher values are better for every reported metric. Higher Noisiness/NOISE/BAK scores mean a better predicted background, while higher Loudness/LOUD scores mean more suitable loudness rather than a louder signal. We keep the predictor families separate because they differ in training and calibration, and do not average their outputs into one MOS. Their dimensional scores are diagnostic predictions, not direct measurements of physical artifacts.

  For the full-band outputs, the reference DNSMOS implementation resamples audio to 16 kHz, while SIGMOS operates at 48 kHz~\cite{dnsmosrepo,sigmosimplementation}. NISQA v2 supports variable sampling rates~\cite{nisqa}.
  
  Before computing reference-based metrics, we compensate for each system's output delay: 16 ms for Enhance (128 samples at 8 kHz or 768 at 48 kHz), 20 ms for RNNoise 0.2 (160 samples for its downsampled 8 kHz output or 960 at 48 kHz), and 40 ms for \dpdf\ (1920 samples at 48 kHz). Noisy baselines need no compensation. This aligns enhanced signals with their clean references and is separate from application buffering or end-to-end latency.
  
  Quality results are means across the benchmark files. Tables~\ref{tab:quality}, \ref{tab:quality8}, and~\ref{tab:quality-cost} show these means, with bold values marking the highest scores at the displayed precision and ties marked equally. We use two decimal places and scaled units where needed. Bold type describes the results rather than statistical significance. We consider the predictors together, since a small numerical gap alone does not establish perceptual equivalence.
  
  We evaluate RNNoise 0.2 and the released \texttt{dpdfnet2\_48khz\_hr} checkpoint on the same EARS--WHAM v2 files. The full-band comparison includes ESTOI, PESQ-WB, SI-SDR, and all three MOS-prediction families. Table~\ref{tab:quality-cost} places the three overall predictions alongside weighted model arithmetic. This compares the deployed checkpoints without isolating the effects of their training procedures.

  \subsection{Model arithmetic and counting scope}
  We count one scalar multiply--accumulate as one MAC and report millions of MAC per second of input audio (MMAC/audio-s). For ONNX models, we use the executed convolution and matrix-product shapes, including grouped convolutions and input/recurrent GRU or LSTM products. We exclude biases, activations, normalization, elementwise recurrent gates, quantization overhead, and external DSP. An INT8 or FP32 multiply--accumulate counts as one MAC, although its runtime cost depends on the precision. We scale the per-step count by the actual internal hop: 62.50 steps/s for the SDK models and 100 steps/s for DPDFNet2.

  The ONNX profiler records executed weighted operations and checks that instrumentation leaves model outputs unchanged. Profiling uses ONNX 1.23.2 and ONNX Runtime 1.23.2, while timing-test versions are given separately. The \enEight\ graph requires 397,688 weighted MAC per 16 ms hop, or 24.86 MMAC/audio-s. The total for \enHybrid\ is 1319.83 MMAC/audio-s, including forward Mel projections within the graph. All comparisons in Table~\ref{tab:quality-cost} include weighted projections within the ONNX graphs.

  For the released \dpdf\ graph, this convention gives 22,839,680 MAC per 10 ms hop, or 2283.97 MMAC/audio-s. We obtain the same count for the sherpa-onnx release asset and the official Ceva export. Published profiles list 2.42 GMAC~\cite{dpdfnetrepo,sherpaenhancement}, which we report separately from the common weighted-operator count used here.

  We count RNNoise operations using the regular model in release 0.2 (model identifier \texttt{0b50c45}). Its dense affine layers require 2,839,296 MAC per 10 ms frame, while the shipped implementation executes retained $8\times4$ weight blocks and separate recurrent diagonals~\cite{rnnoisekernels}. Counting these products gives 1,073,280 MAC per active frame, or 107.33 MMAC/audio-s, compared with 283.93 for the dense equivalent.

  The sparse count includes every product in a retained block, even if an individual stored value is zero. It covers neural weighted products but excludes FFT, pitch analysis, feature extraction, and other DSP. The unmodified library may skip neural inference on detected silence~\cite{rnnoiseapi}; our arithmetic counts assume active frames, while capacity measurements keep the library's standard behavior on the common recording.

  These counts use an explicit rule, but they do not give full pipeline arithmetic or measured CPU operations. For example, they include Mel projections within the SDK graphs, while RNNoise feature extraction is outside its neural count. The complete-path timing below includes each system's signal-processing stages, so we assess speed from measurements rather than MAC ratios alone.

  \subsection{Processing-time measures}
  For an input of duration $D$, SDK real-time factor is defined as
  \begin{equation}
   \mathrm{RTF}_{\mathrm{SDK}}=\frac{\sum_i t_i+t_{\mathrm{final}}}{D},
  \end{equation}
  where $t_i$ is the processing time of call $i$ and $t_{\mathrm{final}}$ is finalization time. Inference RTF uses total neural inference time as the numerator. Loading, decoding, and initialization are measured separately. Processing-time percentiles describe the duration of SDK calls, not end-to-end audio latency.
  
  \subsection{Execution environment}
  The tests ran on an Intel Core Ultra 5 125H laptop with 32 GB of installed RAM. The Linux environment exposed 18 logical CPUs and approximately 15 GiB of RAM. It used Ubuntu 24.04.5 LTS under WSL2 (WSL version 3.0.3.0, kernel \texttt{6.18.54.1-microsoft-standard-WSL2}) on Windows 11 build 26300.9711, with Python 3.12.3 for the benchmark driver. The measurements therefore describe this WSL2 environment rather than Linux running directly on the hardware.
  
  \subsection{Concurrent CLI processes}
  To study how chunk size and process count affect total throughput, we tested three models with 16, 64, and 128 ms chunks and 1, 2, 4, or 8 concurrent CLI processes, giving 36 configurations.
  
  The C++ CLI loads a WAV file into memory once and creates a stream context before a 60 s benchmark interval. It then processes the audio continuously, returning to the start at the end of the file while keeping the same context. Output is neither accumulated nor saved. Chunk size sets the amount of audio passed to the SDK per call, while the spectral-analysis hop remains 16 ms. A Python driver starts each group of CLI processes together and waits for them to finish before the next configuration. Each configuration was measured once.
  
  Each worker measures SDK time per chunk and records accepted audio samples, total processing time, RTF, and call-time percentiles. Aggregate throughput is the total accepted audio duration divided by the group's elapsed wall time, including process startup and termination, whereas worker RTF uses processing time alone. The benchmark also records the maximum worker RTF and the largest of the workers' individual p99 call times. The reported p99 is therefore not a percentile computed from all calls, and neither it nor worker RTF gives end-to-end audio latency.
  
  The driver samples process RSS through \texttt{/proc} every 200 ms and records the largest observed sum across CLI processes. This is the sampled peak of the group sum, which can differ from the sum of individual process peaks. It includes resident input audio and process overhead, may count shared pages more than once, and may miss peaks between samples. Per-process logs and launch parameters are saved with the aggregate results.
  
  \subsection{Shared-model stream capacity}
  In a separate experiment, all streams run in one process and share model resources, while each has its own processing context and OS worker thread. SDK and sherpa-onnx contexts share an ONNX session; RNNoise contexts share weights and use native kernels. Workers process a repeated 20 s WAV as fast as possible with 16 ms input chunks. Each measurement lasts 10 s, with model loading and context preparation completed beforehand. Workers are pinned round-robin across available CPUs, and CPU order is shuffled before each measurement. We allow 30 s of rest between measurements and 3 min between full searches. This measures capacity under the stated affinity policy and complements the separate-process throughput experiment.

  For $P$ concurrent streams, let $D_j$ be the duration in seconds of audio processed by stream $j$ during the common interval $T=10$ s. We use wall-clock RTF,
  \begin{equation}
   R_j^{\mathrm{wall}}=\frac{T}{D_j},\qquad
   R_{\max}^{\mathrm{wall}}=\max_{1\leq j\leq P}R_j^{\mathrm{wall}}.
  \end{equation}
  An attempt passes if $R_{\max}^{\mathrm{wall}}\leq1$, meaning that every stream processes at least 10 s of audio within the common 10 s window. Faster streams cannot compensate for a slower one. This measure includes waiting and scheduling within the interval, whereas SDK-call RTF counts processing calls. We accept a candidate stream count if any of up to three attempts passes, stopping after the first success.

  We perform three complete automated searches per configuration. In each search, we select the successful measurement with the largest stream count and keep the slowest stream's wall RTF from that measurement. We report the mean $\pm$ sample standard deviation of the three counts and, separately, of their three RTFs. Since these RTFs come from selected passing measurements, they are not independent confirmation runs. The retry rule finds demonstrated passing counts but does not guarantee repeated success at that count or that each chunk meets its deadline.

  The external capacity reference is sherpa-onnx 2.0.0 with \texttt{dpdfnet2\_48khz\_hr} at 48 kHz. We adapt it to share one model and ONNX session across independent stream contexts without changing the neural network or signal-processing algorithms. It uses the same input WAV, external chunk size, measurement window, acceptance rule, worker placement, and rest intervals. Inference uses one internal thread. Sherpa keeps its 10 ms internal hop and handles the 16 ms input chunks through its standard adapter, while SDK models use a 16 ms hop. Equal external chunks therefore do not imply equal numbers of inference calls.

  RNNoise 0.2 uses the unmodified library and regular 48 kHz model, with shared weights and an independent \texttt{DenoiseState} and worker for each stream. We use the same 10 s windows, repeated 20 s input, 16 ms external chunks, affinity policy, retries, and rest intervals. Loading and warm-up finish before timing. Internally, RNNoise processes 480-sample (10 ms) frames, so the adapter retains residual samples across 768-sample input chunks. Its API uses float samples on the 16-bit PCM scale~\cite{rnnoiseapi}. Timing covers the full library path, including FFT/iFFT, features, and neural inference, and describes the native build rather than the WebAssembly wrapper.

  Before timing, Sherpa warms each context with three internal silence hops and resets its state, while \sdk\ warms the neural network when loading the model. We keep each system's runtime version and settings: ONNX Runtime 1.28.2 for sherpa-onnx and 1.20.1 for \sdk. These versions match sherpa's Linux x86-64 dependency configuration~\cite{sherpaort} and the SDK dependency lockfile. We time full DSP and neural processing in all three systems. The results compare deployed configurations, so differences cannot be assigned to architecture, ONNX Runtime version, or one implementation feature alone.

  \subsection{Single-stream CLI and Web measurements}
  To compare CLI and Web processing costs and the effect of chunk size, we used the same 30-minute mono recording in both environments. The 48 kHz models received the original 48 kHz WAV, while the 8 kHz model received a copy resampled to 8 kHz beforehand. This input preparation was outside the timed processing. Each model was tested with 16, 32, 64, and 128 ms chunks and three independent repetitions per model/chunk/environment configuration, giving 36 native and 36 browser runs. Runs processed the whole file as fast as possible without saving output. These finite-file tests are separate from the 60 s concurrent-process experiment.
  
  Native execution used ONNX Runtime in Docker under Ubuntu 24.04.5 LTS in WSL2, with a separate CLI process for each repetition. Browser execution used headless Chromium 151.0.7922.34 in the same Linux environment and a Release build of the Web SDK with ONNX Runtime in WebAssembly. Each repetition created a new browser instance, profile, model, and stream context. Processing ran in a Worker without playback, using one inference-engine thread. Native and browser runs used the same model-specific input sampling rates described above.
  
  We timed neural inference and complete processing calls separately. Native SDK calls include spectral transforms, model execution, reconstruction, and resampling where configured; browser calls also include copying across the JavaScript/WebAssembly boundary. The advance preparation of the 8 kHz input is excluded. Total SDK time includes finalization. SDK RTF divides this total by input audio duration, while inference RTF uses inference time instead. Both exclude loading, decoding, resource acquisition, and initialization.

  We report medians across three runs, and each displayed percentile is the median of the three per-run percentiles rather than a percentile over all calls. Separately computed medians need not sum to the median total time.
  
  We measured CPU use during processing for the native CLI process and, through a Linux cgroup, for all processes of the dedicated browser. CPU percentage is total CPU time divided by elapsed measurement time and multiplied by 100, so 100\% means one fully occupied logical CPU. Native measurements use processing-loop elapsed time, while browser measurements use the cgroup monitoring interval. We sampled peak RSS approximately every 200 ms for the native process and for the sum of browser-process RSS. Browser RSS includes browser infrastructure and input buffers, so it cannot be compared directly with native RSS as a measure of model memory. Neither measurement isolates model weights, and sampling may miss brief peaks.
  
  We use CLI to denote native execution and Web to denote WebAssembly. $R_{\mathrm{inf}}$ and $R_{\mathrm{SDK}}$ denote inference and complete SDK RTF, respectively. $p_{50}$, $p_{95}$, and $p_{99}$ give processing-call times, and $p_{99}^{\mathrm{inf}}$ gives inference-call time, all in milliseconds. For display, RTF columns use $100R$ (percent of audio duration) so that small values remain distinguishable with two decimal places; divide the entries by 100 to recover RTF. Bold values mark the lowest inference and SDK RTF within each model/path group.
  
  For initialization time and resource use, we first take the median of three repetitions at each chunk size and then report the minimum and maximum across the four sizes. These ranges show differences between configurations rather than uncertainty intervals. Native loading and decoding were timed together, while Web file reading, decoding, PCM preparation, and resource acquisition were timed separately. We describe repeatability using the coefficient of variation (CV) of SDK RTF for each model/chunk/environment configuration. CV is the sample standard deviation across three runs divided by their mean.
  
  \section{Results and discussion}
  \label{sec:results}
  We first examine quality at 48 kHz and for 8 kHz outputs, including downsampled RNNoise output. We then relate full-band quality to computation and compare native throughput, shared-model capacity, and CLI/Web processing costs.
  
  \subsection{Quality on EARS--WHAM v2}
  Tables~\ref{tab:quality} and~\ref{tab:quality8} compare the models with noisy input at the corresponding sampling rates. The 48 kHz table includes RNNoise and DPDFNet2, while the 8 kHz table includes RNNoise after downsampling its enhanced full-band output.
  
  \begin{table*}[t]
  \centering
  \small
  \setlength{\tabcolsep}{4pt}
  \caption{Speech quality at 48 kHz on EARS--WHAM v2. Entries are mean scores; all PESQ results use WB mode. Reference-based metrics compensate 16, 20, and 40 ms for Enhance, RNNoise, and DPDFNet2, respectively. Bold marks the highest displayed mean in each row.}
  \label{tab:quality}
  \begin{tabular}{lrrrrr}
  \toprule
  Metric & \shortstack{Noisy\\input} & \shortstack{Enhance 48 kHz\\Discriminative} & \shortstack{Enhance 48 kHz\\Hybrid} & \shortstack{RNNoise\\0.2} & \shortstack{DPDFNet2\\48 kHz HR} \\
  \midrule
  ESTOI & $0.64$ & $0.76$ & $0.77$ & $0.70$ & $\mathbf{0.81}$ \\
  PESQ-WB & $1.23$ & $1.88$ & $1.90$ & $1.53$ & $\mathbf{2.21}$ \\
  SI-SDR (dB) & $5.35$ & $12.62$ & $13.03$ & $9.21$ & $\mathbf{14.69}$ \\
  \midrule
  DNSMOS P.808 & $2.68$ & $3.19$ & $3.44$ & $3.33$ & $\mathbf{3.66}$ \\
  DNSMOS SIG & $2.82$ & $3.13$ & $\mathbf{3.29}$ & $3.11$ & $\mathbf{3.29}$ \\
  DNSMOS BAK & $2.43$ & $3.69$ & $3.88$ & $3.99$ & $\mathbf{4.04}$ \\
  DNSMOS OVRL & $2.15$ & $2.71$ & $2.93$ & $2.81$ & $\mathbf{2.99}$ \\
  \midrule
  SIGMOS OVRL & $1.88$ & $2.27$ & $2.65$ & $2.46$ & $\mathbf{2.96}$ \\
  SIGMOS COL & $3.01$ & $2.99$ & $3.31$ & $3.02$ & $\mathbf{3.44}$ \\
  SIGMOS DISC & $3.65$ & $3.33$ & $3.75$ & $3.43$ & $\mathbf{4.01}$ \\
  SIGMOS LOUD & $2.93$ & $3.16$ & $3.43$ & $3.12$ & $\mathbf{3.48}$ \\
  SIGMOS NOISE & $1.61$ & $3.53$ & $3.75$ & $4.17$ & $\mathbf{4.35}$ \\
  SIGMOS REVERB & $4.13$ & $4.39$ & $4.52$ & $4.38$ & $\mathbf{4.57}$ \\
  SIGMOS SIG & $2.88$ & $2.75$ & $3.17$ & $2.77$ & $\mathbf{3.46}$ \\
  \midrule
  NISQA MOS & $2.17$ & $2.69$ & $3.17$ & $2.17$ & $\mathbf{3.81}$ \\
  NISQA Noisiness & $1.90$ & $3.01$ & $3.33$ & $3.80$ & $\mathbf{4.36}$ \\
  NISQA Discontinuity & $4.04$ & $3.55$ & $3.95$ & $2.92$ & $\mathbf{4.34}$ \\
  NISQA Coloration & $3.18$ & $3.04$ & $3.51$ & $2.69$ & $\mathbf{3.94}$ \\
  NISQA Loudness & $2.74$ & $2.86$ & $3.19$ & $2.88$ & $\mathbf{3.38}$ \\
  \bottomrule
  \end{tabular}
  \end{table*}
  
  \begin{table*}[t]
  \centering
  \small
  \setlength{\tabcolsep}{4pt}
  \caption{Speech quality of 8 kHz outputs on EARS--WHAM v2. Entries are mean scores; all PESQ results use NB mode. Enhance 8 kHz INT8 processes downsampled noisy input; RNNoise 0.2 processes the original 48 kHz input before output downsampling. Bold marks the highest displayed mean; input bandwidths differ.}
  \label{tab:quality8}
  \begin{tabular}{lrrr}
  \toprule
  Metric & \shortstack{Noisy\\8 kHz} & \shortstack{Enhance 8 kHz\\INT8} & \shortstack{RNNoise\\48$\rightarrow$8 kHz} \\
  \midrule
  ESTOI & $0.64$ & $\mathbf{0.72}$ & $0.70$ \\
  PESQ-NB & $1.85$ & $\mathbf{2.28}$ & $2.16$ \\
  SI-SDR (dB) & $5.32$ & $\mathbf{10.99}$ & $9.03$ \\
  \bottomrule
  \end{tabular}
  \end{table*}
  
  Both released 48 kHz models improve reference-based and predicted overall quality over noisy input, but differ in their predicted speech distortion. For \enDisc, better noise-related scores come with lower speech-signal and discontinuity predictions than the noisy baseline. \enHybrid\ improves every reported non-intrusive dimension over \enDisc. These improvements suggest better speech preservation than the discriminative model. This interpretation is based on predictors and does not establish listener preference.

  The difference between the 48 kHz models is larger in predicted perceptual quality than in reference-based reconstruction. ESTOI and PESQ-WB are similar, while \enHybrid\ improves SI-SDR by 0.41 dB and raises DNSMOS P.808, SIGMOS OVRL, and NISQA MOS by 0.25, 0.38, and 0.48, respectively. Its main measured benefit is therefore in predicted speech quality, giving applications another choice when they can afford the extra computation.

  For delivered 8 kHz audio, \enEight\ improves both estimated intelligibility and reconstruction over noisy input. It also exceeds downsampled RNNoise 0.2 in ESTOI, PESQ-NB, and SI-SDR, with a 1.96 dB SI-SDR advantage. The agreement across these metrics supports the model as a useful narrowband option: the benefit is present in both the intelligibility estimate and reference-based signal quality, rather than in one score alone. This comparison concerns delivered 8 kHz audio: \enEight\ receives downsampled input, while RNNoise 0.2 processes the original full-band input before output downsampling. The \enEight\ graph requires 24.86 MMAC/audio-s, compared with 107.33 for RNNoise's sparse neural path. Although this is a 4.32-fold difference, it does not show which resampling pipeline is faster, because the counts exclude external DSP and resampling.

  \subsection{Full-band quality and computation trade-offs}
  Table~\ref{tab:quality-cost} relates the three overall predictions to model arithmetic. \enHybrid\ scores higher than RNNoise 0.2 on all three overall measures and on speech-signal quality, discontinuity, coloration, and loudness. RNNoise 0.2 has better noise-related scores and a much lower computation cost. Even on the same recordings, the choice therefore depends on how an application balances noise suppression, speech preservation, and available computation.

  Compared with \dpdf, the hybrid model is competitive on selected speech-quality dimensions. DNSMOS speech quality is tied at the displayed precision, while the DNSMOS OVRL and SIGMOS loudness gaps are 0.06 and 0.05. Other predictions are less close: the SIGMOS coloration gap is 0.13, and DPDFNet2 has a 0.64 advantage in NISQA MOS. \dpdf\ also has higher reference-based scores, all three overall predictions, and clearly stronger noise-related scores. The hybrid therefore approaches DPDFNet2 on some speech-preservation measures, but these results do not establish overall quality equivalence. We describe second-decimal differences without assuming a fixed threshold for perceptual relevance.

  The computation difference also needs to be considered with the quality results. \enHybrid\ uses 1319.83 MMAC/audio-s versus 2283.97 for \dpdf, or 42.21\% fewer counted weighted operations, giving a ratio of 1.73. It therefore offers lower computation cost with quality close to DPDFNet2 on selected speech dimensions, while DPDFNet2 retains stronger overall predictions. RNNoise has still lower sparse arithmetic, although complete-path timing is needed to compare practical processing costs.

  This balance reflects the model's intended use in videoconferencing on native platforms and in browsers. Enhancement must provide useful speech quality with limited processing delay while leaving resources for video encoding and decoding, including demanding codecs. Lower model cost helps meet these requirements across a wider range of devices, including less powerful ones.

  The Web results in Table~\ref{tab:execution-cost} show processing headroom on the tested host. The higher cost of the sherpa-onnx/\dpdf\ configuration is likely to leave less CPU headroom on weaker devices, which may make continuous low-delay processing more difficult when video workloads compete for resources. This is an expectation based on processing cost, rather than a measured comparison of browser behavior on those devices.

  \begin{table*}[p]
  \centering
  \small
  \setlength{\tabcolsep}{4pt}
  \caption{Predicted quality and weighted model arithmetic at 48 kHz. Quality entries are mean scores on EARS--WHAM v2. ONNX counts include forward Mel projections within the graphs; RNNoise counts include retained blocks and recurrent diagonals (dense equivalent: 283.93 MMAC/audio-s). FFT and other external DSP are excluded. Section~\ref{sec:eval} details the counting rules.}
  \label{tab:quality-cost}
  \begin{tabular}{lrrrr}
  \toprule
  Model & MMAC/audio s & DNSMOS P.808 & SIGMOS OVRL & NISQA MOS \\
  \midrule
  \enHybrid & 1319.83 & $3.44$ & $2.65$ & $3.17$ \\
  RNNoise 0.2 & 107.33 & $3.33$ & $2.46$ & $2.17$ \\
  \dpdf & 2283.97 & $\mathbf{3.66}$ & $\mathbf{2.96}$ & $\mathbf{3.81}$ \\
  \bottomrule
  \end{tabular}
  \end{table*}

  \subsection{Native processing throughput}
  Table~\ref{tab:native-throughput} reports aggregate throughput for concurrent CLI processes.
  
  \begin{table*}[p]
  \centering
  \small
  \caption{Aggregate native throughput (audio s/s).}
  \label{tab:native-throughput}
  \begin{tabular}{llrrrr}
  \toprule
  Model & Chunk (ms) & $P=1$ & $P=2$ & $P=4$ & $P=8$ \\
  \midrule
  \enEight & 16 & 134.00 & 297.10 & 543.90 & \textbf{660.50} \\
   & 64 & 300.00 & 581.50 & 1058.10 & 1405.60 \\
   & 128 & 347.80 & \textbf{802.90} & 1430.40 & 1772.00 \\
  \midrule
  \enDisc & 16 & 17.90 & 26.70 & 41.50 & \textbf{49.30} \\
   & 64 & 34.70 & 62.20 & 108.50 & 142.80 \\
   & 128 & 49.80 & \textbf{91.40} & 158.30 & 212.60 \\
  \midrule
  \enHybrid & 16 & 9.64 & 17.30 & 29.44 & 36.79 \\
   & 64 & 16.43 & 30.56 & 56.45 & 68.12 \\
   & 128 & 18.13 & 34.95 & 65.23 & 76.94 \\
  \bottomrule
  \end{tabular}
  \end{table*}

  Larger chunks can compensate for fewer processes, although the point at which they do so depends on the model. For \enEight\ and \enDisc, two processes with 128 ms chunks outperform eight with 16 ms chunks by factors of 1.22 and 1.85 (bold values in Table~\ref{tab:native-throughput}). For \enHybrid, the corresponding rates are 34.95 and 36.79 audio s/s, so larger chunks nearly compensate for the fourfold difference in process count. With four processes at 128 ms, the hybrid reaches 65.23 audio s/s, or 1.77 times its eight-process 16 ms rate. The larger hybrid therefore has a different balance between frame grouping and additional processes. These measurements do not separate call overhead from backend efficiency, and live applications also need more time to collect larger chunks.
  
  Increasing the process count from four to eight adds 17.95--34.30\% to throughput across the tested configurations, showing that returns fall as more processes are added. For \enHybrid, increasing chunk size from 16 to 128 ms raises throughput by 1.88 times with one process and 2.09 times with eight. Since chunk size and parallelism affect each other, a single-stream result cannot be scaled linearly to predict server capacity. \enEight\ is the fastest released Enhance model in every configuration and provides a separate narrowband choice.
  
  \subsection{Shared-model capacity under a per-stream rate criterion}
  Table~\ref{tab:concurrency-boundary} reports the largest passing stream counts from three searches and their corresponding slowest-stream wall RTFs. Both external references use shared weights and independent stream states, and the adapted sherpa-onnx reference also shares its model/session.

  \begin{table*}[p]
  \centering
  \small
  \setlength{\tabcolsep}{6pt}
  \caption{Shared-model capacity with 16 ms external input chunks and 10 s measurement windows. Values are mean $\pm$ sample SD across three searches. Wall RTF is taken from the slowest stream in each search's selected successful measurement and displayed as $100R$ (percent of audio duration). For an attempt to pass, every stream's unrounded wall RTF must be $\leq1$.}
  \label{tab:concurrency-boundary}
  \begin{tabular}{lrr}
  \toprule
  Configuration & Greatest passing streams & Selected $100R_{\max}^{\mathrm{wall}}$ \\
  \midrule
  \enEight & $593.33\pm8.33$ & $96.89\pm2.47$ \\
  \midrule
  \enDisc & $151.00\pm1.00$ & $98.23\pm1.39$ \\
  \enHybrid & $56.00\pm1.00$ & $96.71\pm3.02$ \\
  sherpa-onnx / \dpdf\textsuperscript{a} & $17.33\pm0.58$ & $94.55\pm4.45$ \\
  RNNoise 0.2 & $1088.67\pm30.09$ & $99.58\pm0.24$ \\
  \bottomrule
  \end{tabular}
  \par\smallskip
  \begin{minipage}{0.97\textwidth}\footnotesize
  \textsuperscript{a}Shared-model adaptation of sherpa-onnx 2.0.0. RNNoise uses its unmodified native kernels. Runtime versions and internal hops differ as specified in Section~\ref{sec:eval}. The Enhance 8 kHz row is a separate bandwidth condition. RNNoise capacity is measured at 48 kHz without the output-downsampling step used in Table~\ref{tab:quality8}. RTF rounding does not alter the pass criterion, which uses unrounded measurements.
  \end{minipage}
  \end{table*}

  \begin{table*}[p]
  \centering
  \small
  \setlength{\tabcolsep}{4pt}
  \caption{Single-stream processing costs. RTF columns show $100R$ (percent of audio duration): divide by 100 to recover RTF. These are time ratios, not CPU-use percentages. Call-time percentiles are in milliseconds.}
  \label{tab:execution-cost}
  \begin{tabular}{llrrrrrr}
  \toprule
  Model & Path / chunk (ms) & $100R_{\mathrm{inf}}$ & $100R_{\mathrm{SDK}}$ & $p_{50}$ & $p_{95}$ & $p_{99}$ & $p_{99}^{\mathrm{inf}}$ \\
  \midrule
  \enEight & CLI / 16 & 0.47 & 0.65 & 0.07 & 0.23 & 0.55 & 0.37 \\
   & CLI / 32 & 0.33 & 0.52 & 0.11 & 0.37 & 0.82 & 0.50 \\
   & CLI / 64 & 0.23 & 0.44 & 0.22 & 0.57 & 1.06 & 0.60 \\
   & CLI / 128 & \textbf{0.10} & \textbf{0.23} & 0.26 & 0.48 & 0.94 & 0.40 \\
  \addlinespace[3pt]
   & Web / 16 & 1.01 & 1.42 & 0.19 & 0.39 & 0.61 & 0.42 \\
   & Web / 32 & 0.66 & 1.02 & 0.30 & 0.55 & 0.84 & 0.54 \\
   & Web / 64 & 0.45 & 0.79 & 0.47 & 0.80 & 1.20 & 0.68 \\
   & Web / 128 & \textbf{0.35} & \textbf{0.69} & 0.82 & 1.34 & 1.96 & 1.03 \\
  \midrule
  \enDisc & CLI / 16 & 3.55 & 3.97 & 0.56 & 1.09 & 1.59 & 1.42 \\
   & CLI / 32 & 3.19 & 3.53 & 1.03 & 1.79 & 2.49 & 2.27 \\
   & CLI / 64 & 2.13 & 2.44 & 1.44 & 2.46 & 3.30 & 2.95 \\
   & CLI / 128 & \textbf{1.61} & \textbf{1.90} & 2.25 & 3.78 & 4.84 & 4.23 \\
  \addlinespace[3pt]
   & Web / 16 & 12.24 & 12.96 & 1.95 & 2.98 & 3.90 & 3.70 \\
   & Web / 32 & 8.29 & 8.87 & 2.67 & 4.14 & 5.20 & 4.91 \\
   & Web / 64 & 6.19 & 6.74 & 4.05 & 5.98 & 7.09 & 6.65 \\
   & Web / 128 & \textbf{4.75} & \textbf{5.23} & 6.32 & 8.81 & 10.11 & 9.30 \\
  \midrule
  \enHybrid & CLI / 16 & 9.56 & 9.89 & 1.52 & 1.95 & 2.84 & 2.74 \\
   & CLI / 32 & 7.27 & 7.52 & 2.33 & 2.91 & 3.94 & 3.83 \\
   & CLI / 64 & 5.79 & 6.00 & 3.72 & 4.57 & 5.60 & 5.45 \\
   & CLI / 128 & \textbf{5.34} & \textbf{5.56} & 7.00 & 8.18 & 8.98 & 8.68 \\
  \addlinespace[3pt]
   & Web / 16 & 23.70 & 24.38 & 3.81 & 4.78 & 5.81 & 5.65 \\
   & Web / 32 & 17.20 & 17.66 & 5.52 & 6.83 & 7.59 & 7.41 \\
   & Web / 64 & 14.23 & 14.60 & 9.15 & 10.87 & 11.83 & 11.54 \\
   & Web / 128 & \textbf{12.76} & \textbf{13.09} & 16.57 & 18.62 & 19.94 & 19.41 \\
  \bottomrule
  \end{tabular}
  \end{table*}

  \begin{table*}[p]
  \centering
  \small
  \caption{Initialization time and resource use.}
  \label{tab:execution-resources}
  \begin{tabular}{lrrr}
  \toprule
  Model / path & Initialization (ms) & CPU (\%) & RSS (MiB) \\
  \midrule
  \enEight{} / CLI & 125.75--256.37 & 115.69--156.28 & 443.64--572.26 \\
  \enEight{} / Web & 150.95--160.66 & 111.07--115.25 & 1794.09--1842.16 \\
  \enDisc{} / CLI & 87.43--111.91 & 105.71--106.50 & 448.77--548.29 \\
  \enDisc{} / Web & 268.39--296.08 & 105.64--107.15 & 1772.93--1845.41 \\
  \enHybrid{} / CLI & 87.72--105.05 & 100.00--103.12 & 591.96--640.96 \\
  \enHybrid{} / Web & 302.24--311.38 & 98.67--101.76 & 1797.81--1898.26 \\
  \bottomrule
  \end{tabular}
  \end{table*}

  The shared-model results show the costs of bandwidth and quality selection. \enEight\ supports 593.33 passing streams on average, compared with 151.00 for \enDisc\ and 56.00 for \enHybrid. Its capacity is therefore 3.93 times the discriminative capacity and 10.60 times the hybrid capacity, although it operates at a different bandwidth. Within the 48 kHz pair, the hybrid's quality gains come with a 62.91\% reduction in passing stream count. Model selection consequently affects service capacity as well as the quality of a single stream.

  Compared with the shared-model sherpa-onnx/\dpdf\ deployment, \enDisc\ and \enHybrid\ support 8.71 and 3.23 times as many passing streams. The hybrid offers a useful balance: some speech-quality predictions are close to DPDFNet2, while its measured capacity is higher. Its 3.23-fold capacity ratio exceeds the 1.73-fold weighted-arithmetic ratio, showing why arithmetic alone cannot describe deployment efficiency. These ratios compare complete configurations with different networks, DSP, internal hops, and runtime versions; they cannot be interpreted as an SDK or inference-engine speedup alone.

  RNNoise 0.2 has the highest measured capacity at $1088.67\pm30.09$ streams, or 7.21 and 19.44 times the counts of \enDisc\ and \enHybrid. Its native sparse execution and combined DSP/neural design provide a strong low-cost reference. However, the quality results show why throughput alone is insufficient for model selection. All measured systems share model resources and keep stream states independent, so model sharing by itself cannot explain this ranking.

  Taken together, quality and capacity results suggest three choices. RNNoise 0.2 supports the most passing streams, \enHybrid\ uses more computation for stronger predicted speech quality, and \dpdf\ has the strongest overall predictions with the lowest measured capacity. These choices apply to the stated benchmark and timing protocol. Since the search uses short windows, retries, and a fixed affinity policy, its passing rates do not establish long-running service limits or per-chunk deadline guarantees.

  \subsection{Native and browser processing costs}
  Tables~\ref{tab:execution-cost} and~\ref{tab:execution-resources} report single-stream processing costs, initialization time, and resource use.

  Larger chunks reduce inference and full SDK RTF for every model in both CLI and Web execution, although the median call takes longer because it processes more audio. Fewer calls may reduce fixed overhead, while larger inputs may improve backend efficiency. Inference RTF also falls, which means that the benefit comes from more than the application wrapper. Larger chunks are therefore useful for throughput, provided that a live application can allow the extra input-collection time.
  
  Web processing is slower than native CLI processing in every matched configuration, taking 1.79--3.27 times as much SDK time. This difference includes backend and wrapper costs as well as WebAssembly execution. All tested configurations still process audio faster than real time, and the model ranking is the same in both paths. Applications therefore face a similar choice between speed and model complexity in either environment. At 128 ms, inference for the 8 kHz INT8 model takes about half of total processing time or less, so further improvements must also address spectral processing, reconstruction, and data handling.
  
  Resource use does not follow processing speed directly. Web RSS is similar across models with different inference costs because it also includes the loaded recording and runtime overhead. Total application memory therefore cannot be estimated from model size alone. The faster model does not always have the lowest CPU percentage or initialization time either. Since processing runs as fast as possible, CPU percentage describes resource use during that work rather than the load needed for a live stream.
  
  Most configurations vary little between repetitions, although native discriminative processing at 16 ms has a CV of about 27\%. Its RTF ranges from 0.04 to 0.06, with every run still faster than real time. Comparisons involving that configuration are therefore less precise. The measurements support the observed processing-cost trends, but do not establish playback continuity or end-to-end latency.

  Our applications use the SDK on devices with different processing capabilities and microphone quality. Integration testing included concurrent video streams, reflecting the videoconferencing workloads that shaped the design. The released models balance speech quality against the processing budget needed to keep audio running alongside other tasks. These application checks provide practical experience, while the file-processing benchmarks quantify headroom on the tested host rather than playback interruptions or comparative browser stability under video workloads.
  
  \section{Conclusion}
  We presented \sdk, an open-source SDK that combines neural denoising and configurable audio processing in reusable enhancement channels. The release includes an 8 kHz INT8 model and discriminative and hybrid 48 kHz FP32 models, together with selectable level-control and dynamics blocks, independent input/output sample-rate conversion, and SpeexDSP echo cancellation on supported interfaces. Applications can use normalization alone, denoising alone, or both. Native interfaces, Android and iOS SDKs, and Web execution use one C++ core, with ONNX Runtime and TensorFlow Lite support as listed in the platform matrix.

  Our system contribution combines these configurable channels with shared model resources, independent stream contexts, and the ability to process consecutive frames in one inference call. All tested native and Web configurations process files faster than real time. With 16 ms chunks, median complete SDK processing time is 0.65--9.89\% of audio duration in native execution and 1.42--24.38\% in Web execution across the three models. The larger hybrid model approaches \dpdf\ on selected speech-quality predictions while supporting 3.23 times its shared-model passing stream count, although DPDFNet2 retains stronger overall-quality predictions. These results support a configurable approach in which applications choose quality, bandwidth, and computation together. Chunk-size effects and full-path costs show that the choice also depends on how audio is processed. All reported SDK timing uses ONNX Runtime.

  The model trade-offs come from practical videoconferencing requirements on devices with different hardware capabilities and microphone quality. Enhancement has to leave room for demanding video codecs while limiting audio delay and processing artifacts. Internal application tests with concurrent video streams informed these choices and complement the quantitative experiments, which cover one host environment and one external quality benchmark. Regression tests check consistency across chunk sizes, while runtime measurements show processing headroom rather than per-chunk deadline guarantees. Human listening preferences, long-duration service limits, and quantitative robustness across devices and competing workloads remain to be studied. The public release provides the SDK and model configurations for further work on these deployment questions.

  \end{document}